\documentclass[runningheads]{llncs}
\usepackage{graphicx}

\usepackage{algorithm}
\usepackage{algpseudocode}
\usepackage{amsfonts}
\usepackage{amsmath}
\usepackage{booktabs}
\usepackage{enumitem}
\usepackage{graphicx}
\usepackage{indentfirst} 
\usepackage{microtype}
\usepackage{multicol}
\usepackage{multirow}
\usepackage{latexsym}
\usepackage{lipsum}
\usepackage{listings}
\usepackage{pdfpages}
\usepackage{hyperref}
\usepackage{tabularx}
\usepackage{tikz}
\usepackage{xcolor}
\usepackage{xspace}
\usepackage{subfigure}
\usepackage{subcaption}
\usepackage{mdframed} 
\usepackage{array} 
\usepackage{amsmath}
\usepackage{cite}

\usepackage{amsmath}
\usepackage{algorithm}
\usepackage{algpseudocode}
\usepackage{amsfonts}
\usepackage{booktabs}
\usepackage{enumitem}
\usepackage{indentfirst} 
\usepackage{microtype}
\usepackage{multicol}
\usepackage{multirow}
\usepackage{latexsym}
\usepackage{lipsum}
\usepackage{listings}
\usepackage{pdfpages}
\usepackage{hyperref}
\usepackage{ulem} 
\usepackage{cancel} 
\usepackage{tabularx}
\usepackage{tikz}
\usepackage{xcolor}
\usepackage{xspace}
\usepackage{subcaption}
\usepackage{mdframed} 
\usepackage{cleveref}
\usepackage{xparse} 
\usepackage{arydshln}
\usepackage{wrapfig}

\newcolumntype{?}{!{\vrule width 1pt}}

\newcommand{\sysname}{\textsc{NoiSec}\xspace}

\usepackage{tikz}
\usepackage{collcell}

\usepackage{etoolbox}
\newtoggle{inTableHeader}
\toggletrue{inTableHeader}

\newcommand*{\StartTableHeader}{\global\toggletrue{inTableHeader}}%
\newcommand*{\EndTableHeader}{\global\togglefalse{inTableHeader}}%

\let\OldTabular\tabular%
\let\OldEndTabular\endtabular%
\renewenvironment{tabular}{\StartTableHeader\OldTabular}{\OldEndTabular\StartTableHeader}%

\newcommand*{\MinNumber}{0.50}%
\newcommand*{\MidNumber}{0.75} %
\newcommand*{\MaxNumber}{1.0}%

\definecolor{lightred}{rgb}{1.0, 0.8, 0.8}   
\definecolor{lightgreen}{rgb}{0.0, 1.0, 0.0} 
\newcommand{\ApplyGradient}[1]{%
  \iftoggle{inTableHeader}{#1}{
    \ifdim #1 pt > \MidNumber pt
        \pgfmathsetmacro{\PercentColor}{max(min(100.0*(#1 - \MidNumber)/(\MaxNumber-\MidNumber),100.0),0.00)} %
        \hspace{-0.33em}\colorbox{lightgreen!\PercentColor!yellow}{#1}
    \else
        \pgfmathsetmacro{\PercentColor}{max(min(100.0*(\MidNumber - #1)/(\MidNumber-\MinNumber),100.0),0.00)} %
        \hspace{-0.33em}\colorbox{lightred!\PercentColor!yellow}{#1}
    \fi
  }}

\newcolumntype{R}{>{\collectcell\ApplyGradient}c<{\endcollectcell}}

\definecolor{myblue}{rgb}{0.0, 0.0, 1.0}
\definecolor{mydarkgreen}{rgb}{0.0, 0.5, 0.0}

\newcolumntype{?}{!{\vrule width 1pt}}

\begin{document}
\title{Let the Noise Speak:\\ Harnessing Noise for a Unified Defense Against Adversarial and Backdoor Attacks}
\titlerunning{Let the Noise Speak}

\author{Md Hasan Shahriar\inst{1}\and
Ning Wang\inst{2} \and
Naren Ramakrishnan\inst{1} \and
Y. Thomas Hou\inst{1}\and
Wenjing Lou\inst{1}
}
\authorrunning{M. Shahriar et al.}
%
\institute{Virginia Polytechnic Institute and State University, Blacksburg, VA, USA\\ 
\email{\{hshahriar, naren, thou, wjlou\}@vt.edu}\\\and
University of South Florida, Tampa, FL, USA\\
\email{ningw@usf.edu}}
\maketitle              
\makeatletter
\renewcommand{\@makefnmark}{\hbox{\@textsuperscript{}}}
\makeatother
\footnotetext{A version of this paper has been accepted by ESORICS 2025.}

\begin{abstract}
The exponential adoption of machine learning (ML) is propelling the world into a future of distributed and intelligent automation and data-driven solutions. However, the proliferation of malicious data manipulation attacks against ML, namely adversarial and backdoor attacks, jeopardizes its reliability in safety-critical applications. The existing detection methods are attack-specific and built upon some strong assumptions, limiting them in diverse practical scenarios. Thus, motivated by the need for a more robust, unified, and attack-agnostic defense mechanism, we first investigate the shared traits of adversarial and backdoor attacks. Based on our observation, we propose \sysname, a reconstruction-based intrusion detection system that brings a novel perspective by shifting focus from the reconstructed input to the reconstruction noise itself, which is the foundational root cause of such malicious data alterations. \sysname disentangles the noise from the test input, extracts the underlying features from the noise, and leverages them to recognize systematic malicious manipulation. Our comprehensive evaluation of \sysname demonstrates its high effectiveness across various datasets, including basic objects, natural scenes, traffic signs, medical images, spectrogram-based audio data, and wireless sensing against five state-of-the-art adversarial attacks and three backdoor attacks under challenging evaluation conditions. \sysname demonstrates strong detection performance in both white-box and black-box adversarial attack scenarios, significantly outperforming the closest baseline models, particularly in an adaptive attack setting. We will provide the code for future baseline comparison.
Our code and artifacts are publicly available at \url{https://github.com/shahriar0651/NoiSec}.

\end{abstract}

\section{Introduction}
\label{sec:intro}
The widespread deployment of machine learning (ML) models across diverse distributed and connected environments, including connected and autonomous vehicles, smart cities, health care, and industrial IoT networks, has driven significant technological advancements. At the same time, they are vulnerable to data manipulation attacks
, including adversarial attacks~\cite{szegedy2013intriguing, goodfellow2014explaining, kurakin2018adversarial, madry2017towards, papernot2016limitations, moosavi2017universal, carlini2017towards} and backdoor attacks~\cite{gu2017badnets, turner2019label, nguyen2021wanet}.
While adversarial attacks imperceptibly alter the test data to deceive benignly trained models, backdoor attacks insert subtle triggers in the training data to compromise the inference integrity of the trained model, which is exploited later in the testing phase. Defending against these threats is challenging due to their stealth and sophistication, demanding robust defense strategies.

Various {detection} methods are designed to detect data manipulation attacks, where the fundamental idea is to analyze the existence of malicious components within test input data. Common analysis approaches include feature space inspection~\cite{feinman2017detecting, wang2022manda}, outlier detection~\cite{grosse2017statistical}, input reconstruction~\cite{meng2017magnet}, etc.  
Most of these methods are built upon the assumption that \textit{the malicious inputs will always lead to \emph{noticeable} changes to model prediction}. 
However, such an assumption on attack impact does not always hold, particularly in real-world scenarios. 
Rather, a malicious input can compromise the model's decision only when the perturbation, the target input, and the target model are all synchronized together~\cite{demontis2019adversarial}. 
Conversely, any asynchrony among these components can diminish the effectiveness of the attack, leading to a failure in achieving the desired level of disruption in the final prediction.
For example, during the initial reconnaissance phase, an attacker might choose a very small perturbation to avoid making noticeable changes to the target input, leading to such desynchronized perturbation.
Similarly, in real-world attack scenarios, various natural processes, such as environmental factors, signal processing, sensor encoding, etc., can introduce unforeseen transformations~\cite{kurakin2018adversarial}, leading to desynchronized input. 
Furthermore, in the case of black-box attacks
, the attacker lacks knowledge of the target model and can use a surrogate model as a proxy to launch a transfer attack~\cite{papernot2017practical}. Any subtle differences in the models, such as architectural/parameter-wise disparities, can also disrupt attack synchronization.  
In these desynchronized scenarios, malicious perturbations are less effective and are likely to be overshadowed by the predominant benign features.

Most of the existing detection-based defenses 
struggle against such desynchronized attempts where the malicious features remain latent. We argue that it is also critical to detect both synchronized and desynchronized attempts since it allows the model owner to prepare and react before the attack makes any real cost.  
Therefore, it is imperative to design a detection mechanism that is independent of the attack's ultimate impact, ensuring the ability to identify both types of attacks for a more robust defense.

The existing literature presents two lines of research, each focusing on separate detection mechanisms for adversarial and backdoor attacks, as they stem from distinct vulnerabilities in ML models.
For instance, adversarial samples are identified by higher prediction uncertainty~\cite{feinman2017detecting, wang2022manda}. Backdoor samples, conversely, are detected through higher prediction consistency in the presence of a trigger~\cite{guo2023scale, hou2024ibd}. However, implementing separate defenses for different attacks is impractical and costly, especially in resource-constrained environments. Hence, we aim to bridge the gap in creating a unified defense strategy to counter both adversarial and backdoor attacks simultaneously, which present significant challenges. 

In the search for a unified defense, we observe a common characteristic of adversarial and backdoor attacks: 
they both manipulate testing data by imprinting the non-generalizable features---subtle and stealthy patterns---that are hard for any naive observers to detect but can still induce misclassification in the target model.  
Existing research demonstrated that adversarial attacks leave such malicious footprints in the form of 
random noise~\cite{ilyas2019adversarial} that are perplexing and prone to misclassification. Similarly, the trigger injection in backdoor attacks directly serves this role, with the trigger itself acting as the non-generalizable feature. While the original content is the same for both the benign and malicious inputs, only the accompanying noise (perturbation or trigger) determines the model's response to it.  
Thus, we argue that compared to the defenses that directly analyze the maliciousness of the test 
inputs, disentangling the noise from the original content and analyzing that noise alone enables a more thorough investigation of malicious properties. 

Although the 
disentangled malicious noise may look random to human or rudimentary detectors, we observe that the target model can still analyze its underlying \textit{structure} and reveal the true intent. 
Due to the nature of attack algorithms, adversarial perturbations exhibit gradient alignment with the target model, while backdoor triggers are memorized by the model during backdoor training. 
Therefore, for the same reason, the target model's response to malicious noise will be distinctly different from its response to truly random or benign noise. 
Based on this observation, we propose \sysname, a novel noise-based detector that disentangles the noise from test data to extract the underlying features and use them for recognizing malicious manipulations.  
Our contributions are summarized as follows.

\begin{itemize} 
    \item To overcome the limitations of the existing defense, specifically under practical settings, and bridge the gap between adversarial and backdoor detection, we investigate their shared characteristics and devise a unified detection approach capable of effectively identifying both attacks across white-box and black-box scenarios. 
    \item We propose \sysname, which works beyond those assumptions of the existing methods and utilizes only the noise, the fundamental root cause of such attacks, to detect the existence of malicious data manipulations. 
    \sysname eliminates the requirements of attack data or prior knowledge of training and relies solely on benign data for training and detection, which aligns well with practical settings.
    \item Our comprehensive evaluation of \sysname highlights its high effectiveness across diverse datasets—including basic objects (Fashion MNIST), natural scenes (CIFAR-10), traffic signs (GTSRB), medical images (Med-MNIST), spectrogram-based audio data (Speech Command), and wireless sensing (Activity). \sysname demonstrates resilience against five state-of-the-art adversarial attacks and three backdoor attacks, even under challenging evaluation conditions.
    The evaluation shows that \sysname provides consistently high detection performance with high average AUROC scores in both white-box (0.932) and black-box (0.875) settings across all the adversarial attacks and datasets. 
    Furthermore, \sysname excels with an average AUROC of 0.937 against backdoor attacks on the CIFAR-10 dataset. Moreover, \sysname significantly outperforms the closest baselines in both adversarial and backdoor attack detection. Additionally, \sysname provides high resilience against an adaptive attacker and also shows minimal false positives, highlighting its robustness and practical utility in real-world security applications.
\end{itemize}


\section{Threat Analysis} 
\label{sec:threat-analysis}
This section introduces the adversarial and backdoor attacks, outlines the threat model under consideration, and provides analysis and observations on these attacks. Additionally, two intuitive examples are presented to support these observations, forming the foundation for the proposed defense strategy.

\vspace{-5pt}
\subsection{Data Manipulation Attacks}
\label{sec:attacks}
The malicious data manipulation attacks against ML seek to sabotage the integrity and reliability of the model, particularly by causing incorrect predictions. 
These attacks can manifest in two main forms: adversarial and backdoor attacks. 

\vspace{-10pt}
\subsubsection{{Adversarial Attacks.}} Adversarial attacks occur during the testing phase, where the attacker creates an adversarial example by meticulously crafting subtle adversarial perturbation and adding it to the target input. Let $x^i$ be the i-th original/benign sample, $\delta^i$ be the adversarial perturbation, then the adversarial sample $x_{adv}^i = x^i + \delta^i$. 
Adversarial examples can cause misclassification, even into a target class. The key challenge is to generate $\delta^i$, that lies within a small range \([-\epsilon, +\epsilon]\), making them subtle enough to evade detection. Different adversarial attacks generate $\delta^i$ in different ways. For instance, we consider the gradient-based attacks, including \textit{fast gradient sign method (FGSM)}~\cite{goodfellow2014explaining}, \textit{basic iterative method (BIM)}~\cite{kurakin2018adversarial}, \textit{projected gradient descent (PGD)}~\cite{madry2017towards}, 
\textit{universal adversarial perturbation (UAP)}~\cite{moosavi2017universal}, etc. Moreover, there are optimization-based attacks, such as 
\textit{Carlini \& Wagner (C\&W)}~\cite{carlini2017towards} and query-based black-box attacks, such as  \textit{Square}~\cite{andriushchenko2020square}.

\vspace{-10pt}
\subsubsection{{Backdoor Attacks.}}
While adversarial attacks occur solely during the testing phase, backdoor attacks
, a form of data poisoning attack, are initiated during the training phase and manifest during testing. Specifically, a small trigger pattern is implanted into poisoned training samples to embed a backdoor in the model, which activates upon encountering the same trigger in test samples, potentially leading to misclassification. 
Formally, given the original dataset \( \mathcal{D} = \{(x^i, y^i)\}_{i=1}^{n} \), the poisoned dataset \( \mathcal{D}_{\text{poison}} = \{({x}_{trg}^i, y_{trg}^i)\}_{i \in \mathcal{S}} \) is constructed by adding a trigger $t^i$ to a training samples ${x}^i$ to generate a triggered samples  ${x}_{trg}^i = {x}^i + t^i$. Here,  
 \( \mathcal{S} \subseteq \{1, \dots, n\} \) represents the set of poisoned samples. Different backdoor attacks consider different types/shapes of $t^i$ and manipulate $y_{trg}^i$ differently. The backdoor attacks that 
 we consider are \textit{BadNet}~\cite{gu2017badnets}, \textit{Label-Consistent Attack (LCA)}~\cite{turner2019label}, and \textit{WaNet Attack}~\cite{nguyen2021wanet} attacks. 

\vspace{-5pt}\subsection{Threat Model}
We present the threat model by outlining the attack model, categorizing attack categories and capabilities, and defining defense goals and underlying assumptions.

\vspace{-10pt}
\subsubsection{Attack Model}

Let us assume, in ideal conditions, that the natural input $x_{nat} = x_{org} + \eta_{nat}$ contains original content $x_{org}$ with natural noise $\eta_{nat}$. \textit{\textbf{{Natural noise}} refers to random variations originating from the environment or system, typically modeled as Gaussian noise, i.e., \( \eta_{nat} \sim \mathcal{N}(0, \sigma^2) \).}
In benign but noisy scenarios, the benign input $x_{ben} = x_{org} + \eta_{ben}$, which possesses both the original content $x_{org}$ with some benign noise $\eta_{ben}$. 
\textit{\textbf{Benign noise} is normally as negligible as $\eta_{nat}$ but sometimes can be noticeably high due to environmental conditions or sensor inaccuracies.}
Let $\mathcal{M}$ be the target classifier to be defended, which predicts $x_{ben}$ as class $y_{ben} = arg~max\mathcal{M}(x_{ben})$. If $\mathcal{M}$ is well trained, $y_{ben}$  will mostly be the same as the ground truth $y_{gt}$ (i.e., $y_{ben}\approx y_{gt}$), indicating a high benign accuracy.
On the contrary, the malicious input $x_{mal} = x_{org} + \eta_{mal}$ contains the noise $\eta_{mal}$, which may look like random noise but possesses a systematic and latent malicious structure within it.  
\textit{\textbf{Malicious noise} includes adversarial perturbations ($\eta_{mal} \approx \delta$) or backdoor triggers ($\eta_{mal} \approx t$) designed to compromise the model's integrity and reliability.} The objective of such malicious data manipulation is to change the prediction to $y_{mal} = arg~max\mathcal{M}(x_{mal})$, which is different from $y_{gt}$ (i.e., $y_{mal} \neq y_{gt}$). For practical purposes, we assume that the benign noise retains the same magnitude as the malicious noise but lacks the structural patterns that characterize malicious behavior. Therefore, we generate the benign noise as \(\eta_{\text{ben}} = {randomize}(\eta_{\text{mal}})\).

\vspace{-10pt}
\subsubsection{Attack Categories and Capabilities}

We categorize attacks based on the attacker's capabilities: \textit{Only Testing Phase Attacks} involve crafting adversarial examples by adding malicious noise ($\eta_{mal} \approx \delta$) to exploit vulnerabilities in a deployed benign model. These include white-box attacks, where the attacker has full access to the model's architecture, parameters, and gradients, enabling precise perturbations, and black-box attacks, where the attacker uses a surrogate model or queries the target model iteratively to generate transferable adversarial samples. In contrast, \textit{Both Training and Testing Phase Attacks} allow the attacker to launch backdoor attacks by manipulating training to inject the vulnerabilities into the model. Here, the malicious noise ($\eta_{mal} \approx t$) corresponds to the backdoor trigger. 

\vspace{-10pt}
\subsubsection{Defense Goal and Capabilities}
The defender aims for a testing time defense, and the goal is to detect if any test input has any systematic malicious component. In other words, the ultimate goal is to discriminate between $x_{ben}$ and $x_{mal}$. 
The defender has no information regarding whether the target model contains a backdoor or the specific type or algorithm used for generating the attacks. We assume that the defender has a small representative dataset that contains clean samples spanning all the classes and the computational capacity to train an autoencoder $\mathcal{A}$ on that dataset. We also assume that, along with the final prediction, the defender can also access the feature representation of any given test input. It is further assumed that the attacker cannot compromise the autoencoder or poison the representative dataset, as it is preserved in a secure manner. 


\vspace{-5pt}
\subsection{Attack Similarities}
\label{sec:attack-similarity}
To design a unified defense, we first examine the similarities between adversarial and backdoor attacks. 
Both attacks add malicious noise to the test data---adversarial attacks use subtle perturbations, while backdoor attacks embed triggers. Both rely on the model's poor generalization and sensitivity to such malicious noise. 
The attack similarities lead to some common observations of the malicious noise.
%
\noindent\tikz[baseline=(char.base)]{\node[circle, fill=black, text=white, inner sep=-1pt] (char) {{$O_1$}};} \textbf{Disentanglement of Noise: } 
Malicious noise is imposed on benign samples, making it possible to disentangle them from the original components. For instance, a denoising autoencoder trained solely on benign samples can separate both the benign and malicious noise from the original components.
\noindent\tikz[baseline=(char.base)]{\node[circle, fill=black, text=white, inner sep=-1pt] (char) {{$O_2$}};} \textbf{Target Model's Unique Response to Different Types of Noise:} The model exhibits distinct responses to the malicious noise due to their connection with the model's learned representations.
For example, adversarial perturbations have gradient alignment with the model’s loss function,  
whereas backdoor triggers act as shortcuts by exploiting the model’s learned associations. In both cases, these malicious noise 
leads to systematic activations in the neurons, resulting in high-magnitude features at the representation layers. 
In contrast, benign noise does not have any of these properties, 
hence, they create scattered activations and low-magnitude features that differ significantly from those observed with malicious noise.

\vspace{-5pt}
\subsection{Motivating Examples}
\label{sec:motivating-example}

\begin{figure*}[!t]
    \centering
    \subfigure[Adversarial (BIM) Attack. 
    ]{
    \label{fig:example_case_fashion_a}
    \includegraphics[width=0.475\textwidth]{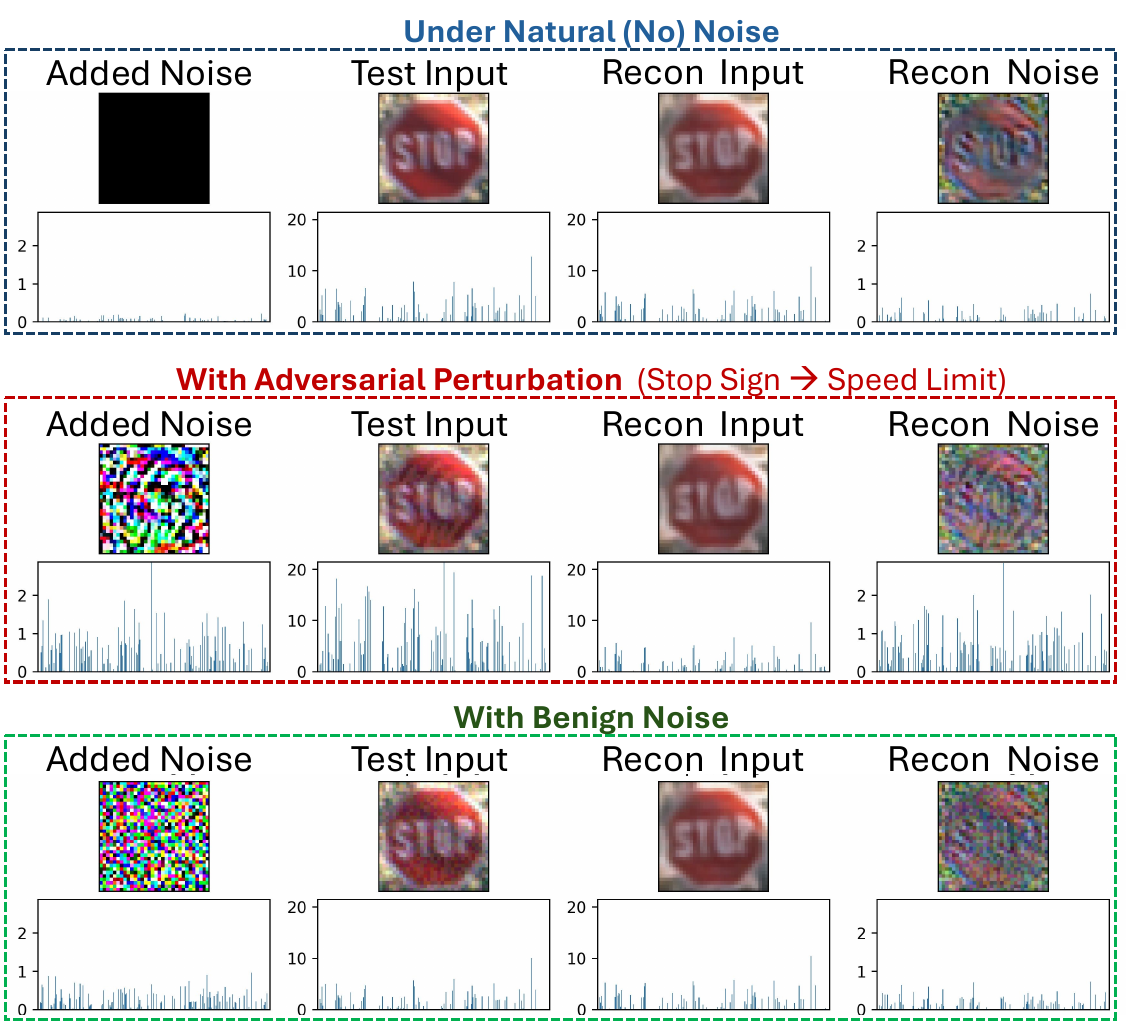}}
    \subfigure[Backdoor (BadNet) Attack. 
    ]{
    \label{fig:example_case_fashion_b}
    \includegraphics[width=0.475\textwidth]{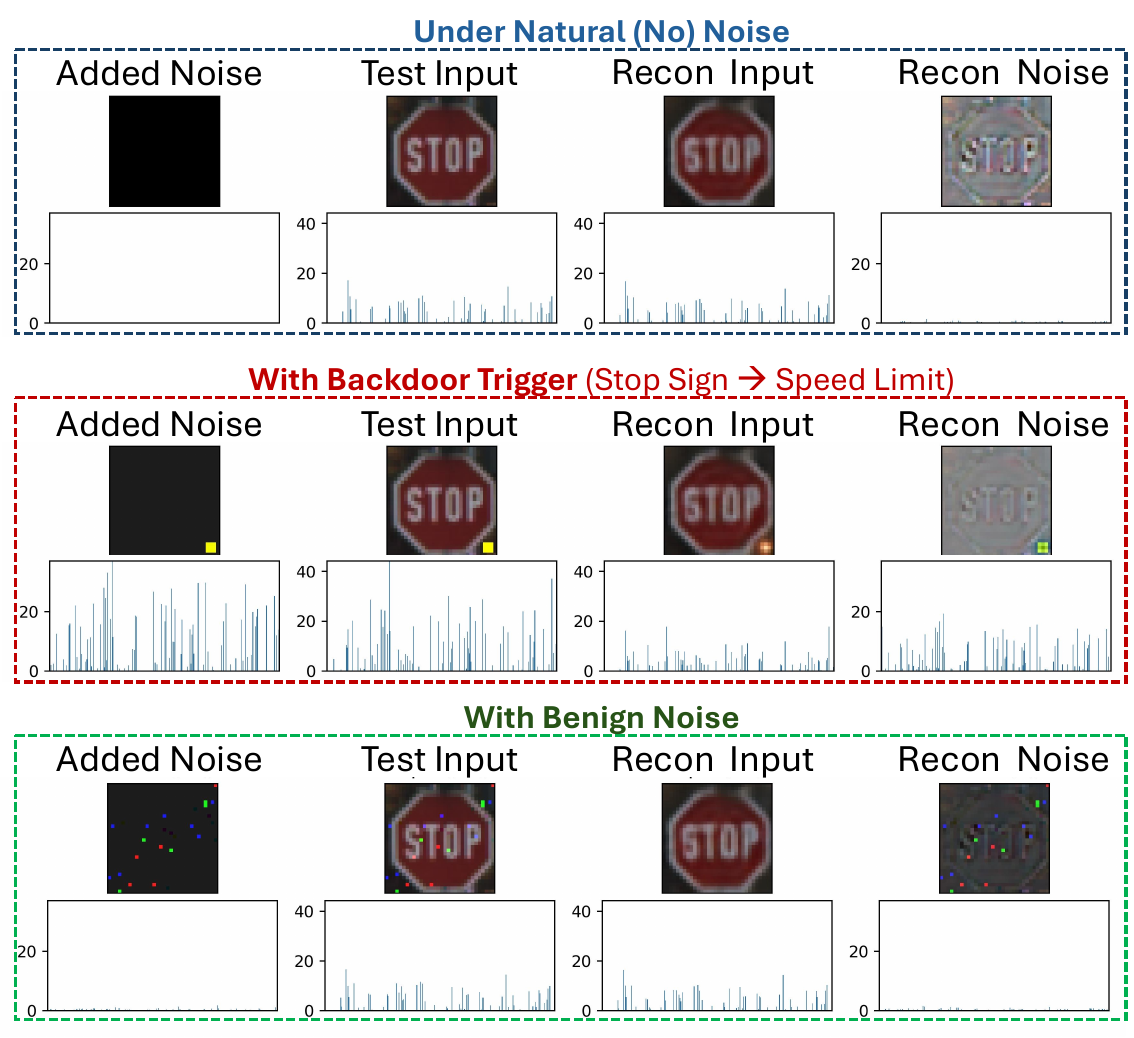}}
\vspace{-7.5pt}
\caption{Effectiveness of using noise to discriminate between malicious (adversarial/backdoor) and benign inputs. The unique feature representations (bar plots at the bottom) of different types of noise (natural, malicious, and benign) indicate the effectiveness of the proposed defense.
}
\vspace{-10pt}
\label{fig:example_case_fashion}
\end{figure*}

We illustrate two motivating examples of adversarial and backdoor attacks on a sample from a traffic sign recognition dataset. We disentangle the noise using a denoising autoencoder (AE) and employ the target classifier to analyze feature representations of different inputs, particularly the noises, at different stages of noise reconstruction.  
Fig.~\ref{fig:example_case_fashion_a} visually demonstrates our observations against a representative adversarial attack, e.g., a BIM attack. The figure consists of three panels, each depicting a different testing scenario under three different types of noises: natural noise, adversarial perturbations, and benign noise. The figure consists of three panels, each depicting a different testing scenario: natural noise, adversarial perturbations, and benign noise (randomized adversarial perturbations). Below each panel, we include the corresponding feature representations extracted by the target classifier model for each input/noise. Here, the first and the fourth columns show added noise and AE-reconstructed noise, respectively, and the two columns in the middle show the test inputs and their reconstructions. It is evident from the leftmost column of the figure that extracted features from the originally added natural noise (top-left) and benign noise (bottom-left) noises do not contain any high-magnitude features. Meanwhile, the feature representation of the adversarial noise (middle row, left column) has significantly different distributions, mostly with higher magnitude components.  This disparity supports \tikz[baseline=(char.base)]{\node[circle, fill=black, text=white, inner sep=-1pt] (char) {{$O_2$}};} underscoring the target classifier's effectiveness in analyzing the noise structure and providing distinctive feature representation that can even visually discriminate between adversarial perturbation and natural/benign noises. 

However, direct access to the originally added noises (leftmost column) is unavailable to the defender, necessitating AE-based noise reconstruction (rightmost column). The feature representations of the reconstructed noises have almost a similar pattern as the original added noises, which supports  \tikz[baseline=(char.base)]{\node[circle, fill=black, text=white, inner sep=-1pt] (char) {{$O_1$}};} and shows the effectiveness of AE-based noise disentanglement. 
Similarly, Fig.~\ref{fig:example_case_fashion_b} visually demonstrates the findings against a representative backdoor attack (e.g., BadNet)  with a 2x2 yellow square-shaped trigger on the bottom right of the test input. 
These findings highlight AE's ability to extract the malicious noise (perturbation or trigger) from the test data and the target model's ability to extract unique features 
to facilitate the detection. 
Such findings support both of our observations in Section~\ref{sec:attack-similarity}, based on which we design our proposed defense \sysname. 

\section{Problem Formulation}
\label{sec:problem}

{The key objective of this study is to develop an effective detector for discriminating between benign and malicious inputs. We innovatively formulate the malicious data detection problem by decomposing input data into two components: original content and noise (either benign or malicious).  
To disentangle noise from the original content, we consider the reconstruction-based approach, particularly using an autoencoder. We categorize such reconstruction-based defenses into two categories: defenses utilizing the input data itself are termed sample-based detection, and defenses utilizing the noise component are termed noise-based detection. Where the ultimate end goal of the sample-based detection is to discriminate between $x_{ben}$ and $x_{mal}$, the noise-based detection considers the detection problem as 
discriminating between $\eta_{ben}$ and $\eta_{mal}$. Both categories of defense have shown effectiveness in detecting malicious patterns. Our solution falls into the noise-based defense category.}

\textbf{Autoencoder-based Reconstruction.~~} Reconstruction-based defense mechanisms have emerged as one of the prominent approaches in detecting and mitigating the impact of malicious data manipulation attacks in ML~\cite{meng2017magnet}.
These methods leverage an autoencoder model $\mathcal{A}$ to reconstruct test input, aiming to disentangle the accompanying noise---whether benign or adversarial---from the natural contents. 
Further analysis of either the reconstruction input or the reconstructed noise indicates the existence of malicious attacks. 
Let the reconstructed natural, benign, and malicious samples be defined as $\hat{x}_{nat}$, $\hat{x}_{ben}$, and $\hat{x}_{mal}$, respectively. If $\mathcal{A}$ is trained sufficiently, the reconstruction will remove any noises, retain only the original contents, and hence: $\hat{x}_{nat} = \mathcal{A}(x_{nat}) \approx x_{org}$, $\hat{x}_{ben} = \mathcal{A}(x_{ben}) \approx x_{org}$, and  $\hat{x}_{mal} = \mathcal{A}(x_{mal}) \approx x_{org}$.
Again, let the reconstruction noise from the natural inputs be $\hat{\eta}_{nat}$, which can be expressed as $\hat{\eta}_{nat} = (x_{nat} - \hat{x}_{nat}) \approx (x_{nat} - x_{org}) =  \eta_{nat}$. Similarly, the reconstruction noise from the benign and malicious can be expressed as $\hat{\eta}_{ben}\approx\eta_{ben}$ and $\hat{\eta}_{mal}\approx\eta_{mal}$, respectively.
Hence, any reconstructed samples approximate only the original content, whereas the reconstruction noises approximate the added noises, either natural, benign, or malicious. 
Such disengagement of noises serves as the fundamental step for any reconstruction-based defense, as it paves the way for further discriminating between benign and malicious inputs.

\section{Our Proposed Defense: \sysname}
\label{sec:detialed_design}

Based on our observation (Section~\ref{sec:attack-similarity}) and motivating examples (Section~\ref{sec:motivating-example}), we propose \sysname, a unified defense against adversarial and backdoor attacks. 

\vspace{-5pt}
\subsection{\sysname Overview}
\label{sec:overview}
\begin{figure*}[!tbh]
    \centering
    \vspace{-10pt}
    \includegraphics[width=0.95\textwidth]{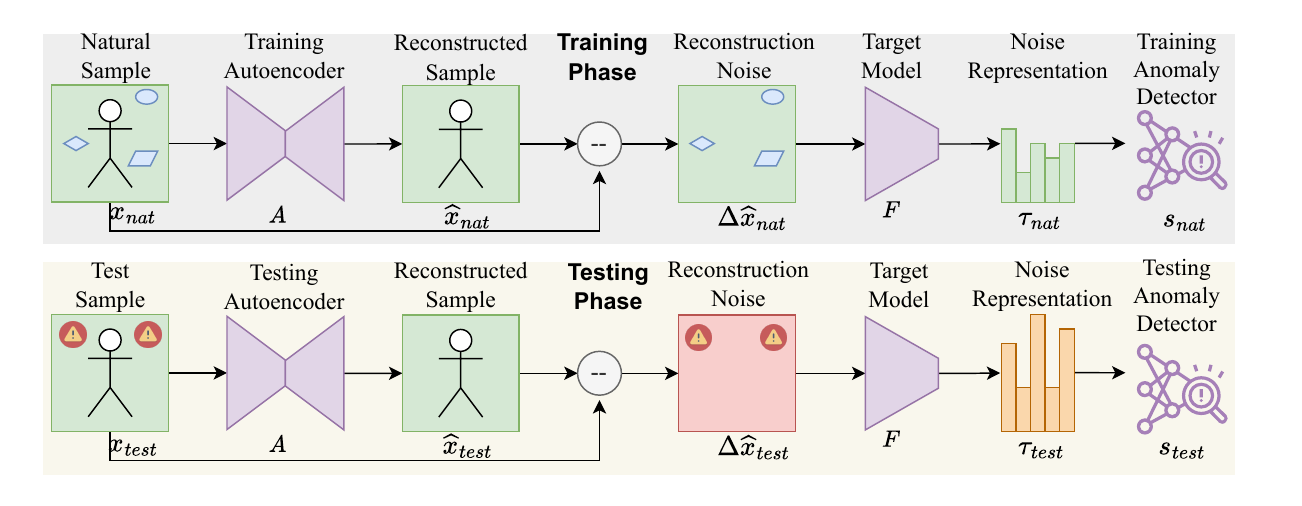}  
    \vspace{-10pt}
    \caption{An overview of the two implementation phases of \sysname.
    }
    \vspace{-15pt}
    \label{fig:noisec}
\end{figure*}

Fig.~\ref{fig:noisec} illustrates the core components and implementation phases of \sysname. It comprises three fundamental components: i) denoising autoencoder, ii) feature extractor (target model), and iii) anomaly detector. Moreover, \sysname has two implementation phases: i) the training phase and ii) the testing phase. 
The training phase, at first, trains the autoencoder (AE) using a representative dataset composed of only natural samples. The AE learns to reconstruct only the original contents and separate the noises from the samples. Later, the trained AE is used to reconstruct all the natural samples and, consequently, calculate the natural reconstruction noises. The natural noises are then fed into the feature extractor (FE) to reduce the dimensionality of the noises and have an effective representation. 

Nonetheless, as natural noises are supposed to have a random structure, all the noise features will exhibit lower magnitudes. Following the acquisition of the low-dimensional noise representation, an anomaly detector (AD) is trained to map the distribution of these natural noise representations and learn the natural pattern or clusters. Finally, \sysname utilizes the trained AD to estimate the anomaly scores of all the natural noise representations and calculates a threshold for future detection. 

During the testing phase, \sysname utilizes the trained AE, FE, and AD, as well as the detection threshold, to check for any malicious manipulation in any test input. As shown in the figure, at the testing phase, the AE reconstructs any incoming test sample (benign or malicious), allowing the estimation of the reconstruction noise. The FE then analyzes such reconstruction noise to have the noise representation. Lastly, the AD analyzes the distribution of this feature vector, contrasts it against the learned natural patterns, and assigns an anomaly score. If the anomaly score exceeds the predefined threshold, \sysname prompts the system to alert for a potential data manipulation attack and take further attack mitigation measures.

\subsection{Technical Details}
This part explains the essential tasks executed sequentially during the training and testing phases of \sysname.

\vspace{-10pt}
\subsubsection{Noise Reconstruction.}

The AE model $\mathcal{A}$ is trained as a denoising AE on the representative dataset to reconstruct the input data while learning to filter out the noise. Upon training of $\mathcal{A}$, the first step involves reconstructing the noise component from the sample using an AE. While in the training phase, these samples are all benign, in the testing phase, they can be both benign and malicious. The process of benign and malicious noise reconstruction $\hat{\eta}_{\text{ben}}$, and $\hat{\eta}_{\text{mal}}$, respectively, is the same for any reconstruction-based defense.
The key novelty of our proposed method mainly lies in the following two steps.

\vspace{-10pt}
\subsubsection{Noise Representation.}

\sysname uses the FE model $\mathcal{F}$ to analyze noise and have  effective noise feature representation.
Notably, $\mathcal{F}$ is essentially the same as the target classifier $\mathcal{M}$. However, instead of getting the confidence vectors at the last layer of $\mathcal{M}$ for noise representation, \sysname considers taking the feature representation at the penultimate layer.
Hence, we separately name this component as $\mathcal{F}$ for clarity, while in implementation, $\mathcal{M}$ itself can be utilized to have this representation.  
Let $\tau_{\text{nat}}$ be the feature representations of the natural reconstructed noises, such that $\tau_{\text{nat}} = \mathcal{F}(\hat{\eta}_{\text{nat}})\approx\mathcal{F}({\eta}_{\text{nat}})$. 
Similarly, let $\tau_{\text{ben}}$ and $\tau_{\text{mal}}$ represent the feature representations of the benign and malicious reconstructed noises, and can be expressed as $\tau_{\text{ben}} = \mathcal{F}(\hat{\eta}_{\text{ben}})\approx\mathcal{F}({\eta}_{\text{ben}})$ and $\tau_{\text{mal}} = \mathcal{F}(\hat{\eta}_{\text{mal}})\approx\mathcal{F}({\eta}_{\text{mal}})$, respectively.




Considering that both $\hat{\eta}_{\text{ben}}$ and $\hat{\eta}_{\text{ben}}$ typically result in feature representations of low magnitude due to the absence of any prominent patterns, $\tau_{\text{ben}}$ is expected to follow the same distribution of $\tau_{\text{nat}}$. Conversely, $\hat{\eta}_{\text{mal}}$, even if with low intensity, is anticipated to activate some specific features, leading to a feature vector of higher magnitude. Hence, the distribution of $\tau_{\text{ben}}$ and $\tau_{\text{nat}}$ are highly similar ($\tau_{\text{ben}} \approx \tau_{\text{nat}}$), while $\tau_{\text{mal}}$ and $\tau_{\text{ben}}$ will have a noticeable difference ($\tau_{\text{mal}} \not\approx \tau_{\text{nat}}$) , which is later also  illustrated in Fig.~\ref{fig:adv_ks_roc_a}. 
Such distinct representations pave the way to the ultimate objective of \sysname, which is to deploy an AD capable of distinguishing between $\tau_{ben}$ and $\tau_{mal}$, thereby identifying potential malicious perturbations.


\vspace{-10pt}
\subsubsection{{Anomaly Detection.}}
Finally, an AD model $\mathcal{D}$ is trained on the natural feature vectors $\tau_{\text{nat}}$ in the training phase and, later in the testing phase, used to discriminate between $\tau_{ben}$ and $\tau_{mal}$. 
Particularly, let the anomaly scores $s_{\text{nat}} = \mathcal{D}(\tau_{\text{nat}})$, $s_{\text{ben}} = \mathcal{D}(\tau_{\text{ben}})$ and $s_{\text{mal}} = \mathcal{D}(\tau_{\text{mal}})$ for natural, benign, and malicious noises representation, respectively. 
Where $ s_{\text{ben}}$ is supposed to have a similar distribution to $s_{nat}$ ($ s_{\text{ben}} \approx s_{\text{nat}}$),  $s_{\text{mal}}$ is assumed to have significantly higher values compared to $s_{\text{ben}}$ ($s_{\text{mal}} >> s_{\text{nat}}$) due to its unforeseen and out of distribution characteristics.
Based on these steps, \sysname effectively discriminates between  $x_{ben}$ and $x_{mal}$, which are evaluated under a wide spectrum of attacks in the following sections. 


\section{Implementation}
\label{sec:implementation}

\vspace{-5pt}
\subsection{Experiment Setup}

\begin{table}[t!]
\centering
\setlength{\tabcolsep}{1pt}
\caption{Comparison of Datasets}
\resizebox{0.85\textwidth}{!}{%
\begin{tabular}{|l|l|l|c|l|}
\hline
\textbf{Dataset} & \textbf{Modality} & \textbf{Input Size}& \textbf{Classes} & \textbf{Description} \\ \hline
\textit{F-MNIST}~\cite{xiao2017fashion} & Image & 28×28x1& 10 & Representations images fashion items.\\ \hline
\textit{CIFAR-10}~\cite{alex2009learning} & Image & 32×32x3& 10 & RGB images of objects, e.g., airplanes.\\ \hline
\textit{GTSRB}~\cite{stallkamp2011german} & Image & 32×32x3& 43 & RGB images of traffic signs.\\ \hline
\textit{SPEECH}~\cite{warden2018speech} & Audio & 64×81x1& 35 & Mel-spectrogram of spoken commands\\ \hline
\textit{Med-MNIST}~\cite{medmnistv2} & X-rays & 64×64x1 & 2 & Chest X-ray images for pediatric pneumonia.\\ \hline
\textit{Activity}~\cite{yousefi2017survey} & Wireless & 500×90x1 & 7 & CSI of wireless sensing of human activities.\\ \hline
\end{tabular}%
}
\vspace{-12.5pt}
\label{tab:dataset_comparison}
\end{table}


We demonstrate \sysname's effectiveness across diverse modalities of datasets, as summarized in Table~\ref{tab:dataset_comparison}. 
We consider various classification models (See Table~\ref{table:model_properties} in Appendix) across different datasets for adversarial attack scenarios. 
It is noteworthy that for all datasets, the target and surrogate models---for white-box and black-box attacks---exhibit varying numbers of channels in their convolutional layers. We use ReLU as the activation function and dropout for regularization. On the other hand, we implement backdoor attacks on the CIFAR-10 dataset using the open-source implementation provided by Backdoorbox~\cite{li2023backdoorbox}, employing the ResNet18 architecture~\cite{he2016deepresnet}. 
Similarly, we consider different autoencoder architectures for different datasets (See Table~\ref{table:autoencoder_properties} in Appendix).
All the models employ 3x3 kernels and ReLU activation functions throughout. We train them as denoising autoencoders, introducing standard Gaussian noise with a standard deviation specified in the table. 
We train both the classifier and the autoencoder using the full training split of their respective datasets. For the AD model, we test various statistical and outlier detection algorithms and find that \textit{Gaussian Mixture Model (GMM)}-based AD performs best. GMM effectively models the data distribution using a combination of Gaussian components~\cite{dempster1977maximum}, capturing both structure and variability in the dataset.

\vspace{-5pt}
\subsection{Evaluation Settings}
\begin{wrapfigure}{r}{0.60\linewidth} 
    \vspace{-15pt}
    \centering
    \includegraphics[width=0.975\linewidth]{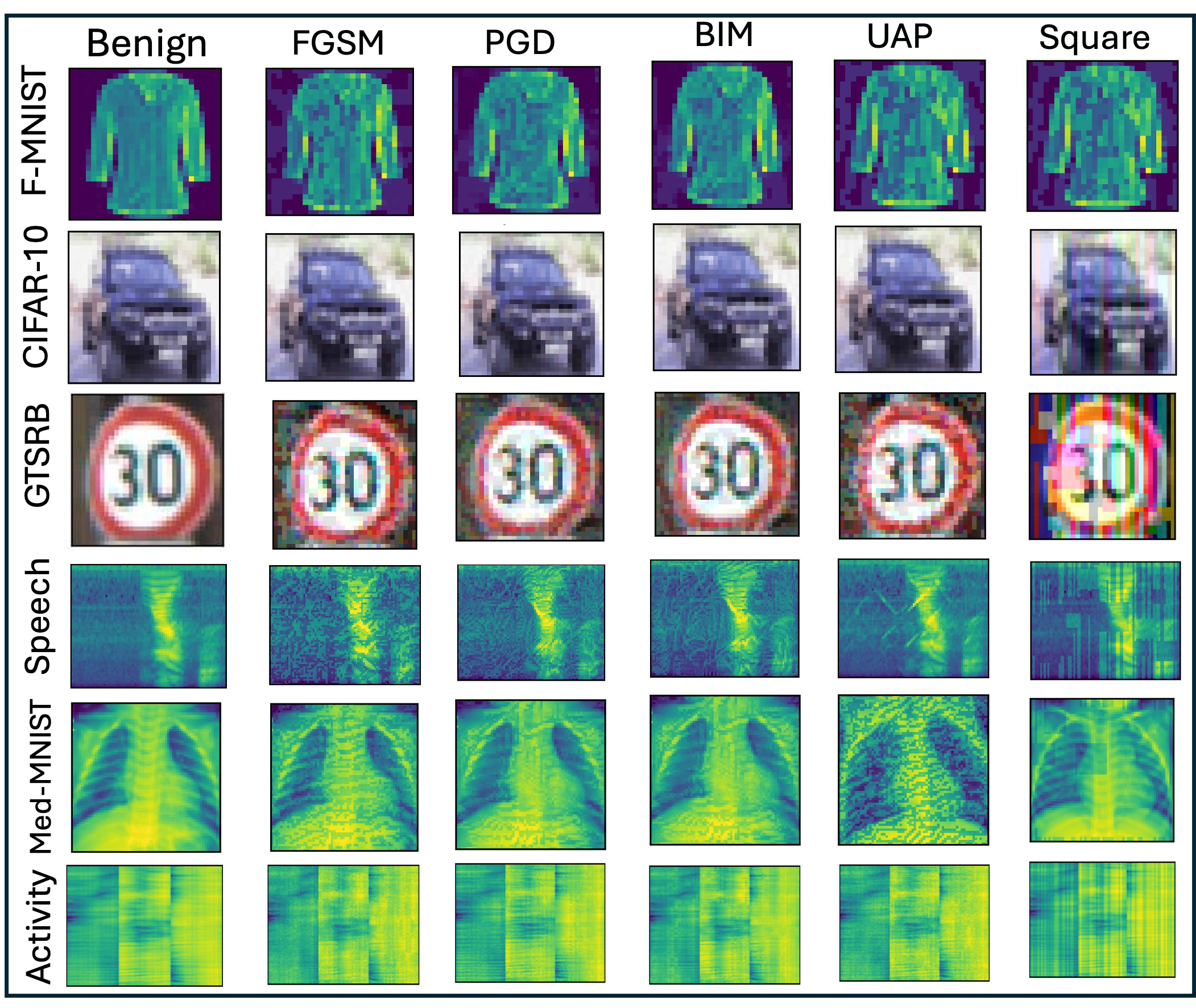}
    \caption{Adversarial examples across attacks.
    }
    \vspace{-15pt}
    \label{fig:attack_grid-adv}
\end{wrapfigure}

We evaluate \sysname against all the attacks mentioned in Section~\ref{sec:attacks}. 
For the adversarial attacks, we generate 500 natural samples by adding Gaussian noise 
for each dataset. 
Subsequently, we generate 100 adversarial samples for each attack using both the target and surrogate models. We randomize the perturbation of each malicious sample and consider them benign samples. Therefore, the benign and malicious sample pairs have the same noise magnitude, but the perturbation structure/pattern differs. This challenging evaluation setting ensures that \sysname only detects malicious inputs but not benign anomalies. Fig.~\ref{fig:attack_grid-adv} shows the samples of adversarial examples across different attacks and datasets. 

We conduct three distinct backdoor attacks on the CIFAR-10 dataset, each with varying poison rates and target labels. We implement BadNet with a poison rate of 5\%, 
using a checkerboard pattern in the bottom-right corner of the image as the trigger. WaNet, on the other hand, 
applies a transformation-based backdoor with a 10\% poison rate, using subtle warping of the input images. Lastly, LCA is implemented with a significantly higher poison rate of 25\%, 
with checkerboard triggers in four corners. 
To evaluate \sysname against these attacks, we generate 1000 backdoor-triggered samples for all three backdoor attacks. As backdoor models are hypersensitive to trigger-like benign noises, we generate another 1000 samples with Gaussian noise as the benign samples. 

\subsection{Software Implementation}
We implement \sysname using Python 3.10. We use PyTorch 
to develop the classifier and the autoencoder. We utilize Torchattacks~\cite{kim2020torchattacks} and Adversarial Robustness Toolbox (ART)~\cite{art2018} libraries for implementing adversarial attacks, Backdoorbox~\cite{li2023backdoorbox} for backdoor models, and we use the PyOD library~\cite{zhao2019pyod} for the AD models. All experiments run on a server equipped with an Intel Core i7-8700K CPU running at 3.70GHz, a GeForce RTX 2080 Ti GPU, and Ubuntu 18.04.3.

\section{Results}
\label{sec:results}

This section analyzes the implementation results of both adversarial and backdoor attacks, as well as the detection performance of \sysname from multiple perspectives, including performance evaluation of the FE, AD, and a comparison with baseline methods, even under an adaptive adversarial setting. 

\begin{wrapfigure}{r}{0.60\textwidth}
    \vspace{-30pt}
    \centering
    \hspace{-15pt}
    \subfigure[KS test]{
    \label{fig:adv_ks_roc_a}
    \includegraphics[width=0.27\textwidth]{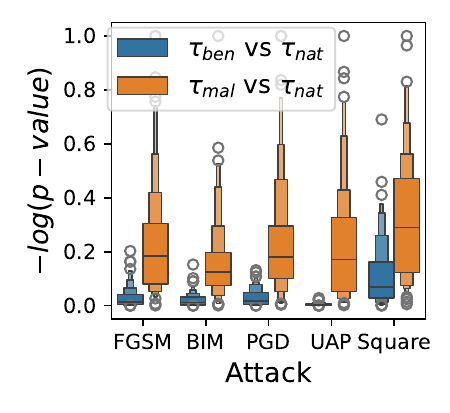}}
    \hspace{-10pt}
    \subfigure[ROC curves]{
    \label{fig:adv_ks_roc_b}
    \includegraphics[width=0.335\textwidth]{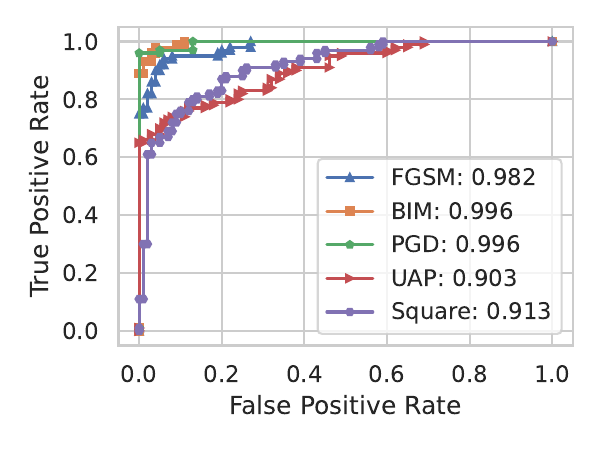}}
    \vspace{-5pt}
    \caption{Performance evaluation of \sysname's FE and AD against different adversarial attacks on CIFAR-10 dataset. (a) 
    KS test results comparing the feature distribution between ($\tau_{ben}$ vs $\tau_{nat}$), and  ($\tau_{mal}$ vs $\tau_{nat}$) for effective feature extraction.
    (b) ROC curves and AUROC scores showing effective anomaly detection.}
        \hspace{-10pt}
    \vspace{-20pt}
\label{fig:adv_ks_roc}
\end{wrapfigure}

\subsection{Effectiveness against Adversarial Attacks}

\subsubsection{Effectiveness of Feature Extractor}
\label{sec:effec_feat_rep_adv}

This part evaluates the efficacy of the target classifier as an FE in capturing critical features indicative of adversarial attacks across various datasets and attack types. We contrast the discrepancies between the feature distributions of reconstructed benign noise ($\tau_{ben}$) and malicious noise ($\tau_{mal}$) by running the Kolmogorov-Smirnov (KS)~\cite{press2007numerical} test on each against the natural noise ($\tau_{nat}$). The {KS} test is a non-parametric test used to assess whether two datasets come from the same distribution or not, where the $-log(p-value)$ of the KS test serves as a measure of the dissimilarity between the two distributions. 
The KS test is employed to compute $-log(p-value)$ for all the features as an indicator for the extent of divergence between each distribution pair.

Fig.~\ref{fig:adv_ks_roc_a} presents the KS test results for different attacks for the CIFAR-10 dataset. 
It is evident that $\tau_{mal}$ exhibits distinct distributions from $\tau_{nat}$, characterized by higher $-log(p-values)$ values for ($\tau_{mal}$ vs $\tau_{nat}$). Conversely, $\tau_{ben}$ and $\tau_{nat}$ generally share similar distributions, indicated 
 by lower $-log(p-values)$ values  from the KS test between ($\tau_{ben}$ vs $\tau_{nat}$). 
This finding further underscores FE's ability to discern structured patterns in adversarial perturbations. 
Overall, this separation is facilitated by effective feature extraction by the target classifiers. 
Such representation enhances the analysis of noise structures and paves the way to more robust anomaly detection. 
Note that we scaled the $-log(p-value)$ values to improve clarity in presentation and comparison. 

\setlength{\tabcolsep}{0.50pt}
\begin{wraptable}{r}{0.58\linewidth} 
\vspace{-22.5pt}
\centering
\caption{AUROC scores of baselines across different attacks and datasets.}
\vspace{-5pt}
\label{tab:auroadversarial_baseline}
\small
\resizebox{0.58\textwidth}{!}{%
\begin{tabular}{|c|c|R|R|R|R|R|R|R|R|R|R|}
\hline 
\multirow{2}{*}{\rotatebox{90}{Data}}
& \multirow{2}{*}{\rotatebox{0}{Defense}} & \multicolumn{5}{c|}{\textbf{White-box}} & \multicolumn{5}{c|}{\textbf{Black-box}} \\ 
\cline{3-12}
& & FGSM & PGD & BIM & UAP & Square &  FGSM & PGD & BIM & UAP & Square \EndTableHeader\\
\hline

\multirow{5}{*}{\rotatebox{90}{F-MNIST}}
 & MagNet & 0.68 & 0.86 & 0.85 & 0.62 & 0.92 & 0.67 & 0.75 & 0.71 & 0.59 & 0.72 \\
 & Artifacts & 0.80 & 0.74 & 0.76 & 0.79 & 0.53 & 0.74 & 0.67 & 0.67 & 0.74 & 0.56 \\
 & Manda & 0.52 & 0.44 & 0.45 & 0.64 & 0.80 & 0.60 & 0.74 & 0.63 & 0.61 & 0.67 \\
 \cline{2-12}
 & \textbf{\sysname} & 0.96 & 0.92 & 0.93 & 0.95 & 0.84 & 0.86 & 0.79 & 0.86 & 0.94 & 0.83 \\
\hline\hline
\multirow{5}{*}{\rotatebox{90}{CIFAR-10}}
 & MagNet & 0.63 & 0.83 & 0.83 & 0.50 & 0.35 & 0.48 & 0.50 & 0.51 & 0.51 & 0.31 \\
 & Artifacts & 0.61 & 0.57 & 0.49 & 0.51 & 0.52 & 0.59 & 0.52 & 0.60 & 0.55 & 0.47 \\
 & Manda & 0.52 & 0.73 & 0.66 & 0.55 & 0.61 & 0.58 & 0.59 & 0.61 & 0.49 & 0.46 \\
 \cline{2-12}
 & \textbf{\sysname} & 0.98 & 1.00 & 1.00 & 0.90 & 0.91 & 0.94 & 0.96 & 0.97 & 0.88 & 0.92 \\
\hline\hline
\multirow{5}{*}{\rotatebox{90}{GTSRB}}
 & MagNet & 0.49 & 0.62 & 0.55 & 0.72 & 0.58 & 0.53 & 0.69 & 0.71 & 0.64 & 0.50 \\
 & Artifacts & 0.43 & 0.72 & 0.86 & 0.53 & 0.58 & 0.48 & 0.53 & 0.56 & 0.54 & 0.58 \\
 & Manda & 0.54 & 0.51 & 0.61 & 0.70 & 0.55 & 0.57 & 0.60 & 0.61 & 0.61 & 0.50 \\
 \cline{2-12}
 & \textbf{\sysname} & 0.89 & 0.88 & 1.00 & 0.84 & 1.00 & 0.83 & 0.87 & 0.90 & 0.70 & 1.00 \\
\hline\hline
\multirow{5}{*}{\rotatebox{90}{Med-MNIST}}
 & MagNet & 0.36 & 0.47 & 0.36 & 0.44 & 0.76 & 0.44 & 0.48 & 0.43 & 0.52 & 0.52 \\
 & Artifacts & 0.56 & 0.61 & 0.53 & 0.56 & 0.76 & 0.63 & 0.54 & 0.62 & 0.53 & 0.66 \\
 & Manda & 0.41 & 0.45 & 0.37 & 0.13 & 0.79 & 0.69 & 0.60 & 0.50 & 0.32 & 0.62 \\
 \cline{2-12}
 & \textbf{\sysname} & 0.90 & 0.83 & 0.98 & 0.99 & 0.91 & 0.74 & 0.67 & 0.79 & 0.80 & 0.89 \\
\hline\hline
\multirow{5}{*}{\rotatebox{90}{Speech}}
 & MagNet & 0.70 & 0.65 & 0.55 & 0.86 & 0.92 & 0.78 & 0.86 & 0.72 & 0.43 & 0.77 \\
 & Artifacts & 0.54 & 0.95 & 0.86 & 0.64 & 0.81 & 0.41 & 0.58 & 0.64 & 0.43 & 0.79 \\
 & Manda & 0.56 & 0.55 & 0.75 & 0.50 & 0.72 & 0.68 & 0.31 & 0.55 & 0.58 & 0.70 \\
 \cline{2-12}
& \textbf{\sysname} & 0.87 & 0.95 & 0.91 & 0.95 & 0.97 & 0.83 & 0.86 & 0.88 & 0.97 & 0.93 \\
\hline\hline
\multirow{5}{*}{\rotatebox{90}{Activity}}
 & MagNet & 0.74 & 0.74 & 0.76 & 0.57 & 0.73 & 0.71 & 0.70 & 0.70 & 0.77 & 0.71 \\
 & Artifacts & 0.68 & 0.70 & 0.76 & 0.51 & 0.67 & 0.67 & 0.70 & 0.64 & 0.61 & 0.74 \\
 & Manda & 0.38 & 0.47 & 0.70 & 0.58 & 0.50 & 0.56 & 0.52 & 0.43 & 0.54 & 0.72 \\
 \cline{2-12}
& \textbf{\sysname} & 0.95 & 0.97 & 0.98 & 0.88 & 0.91 & 0.91 & 0.93 & 0.95 & 0.91 & 0.94 \\

\hline
\end{tabular}
\vspace{-5pt}
}
\end{wraptable}

\vspace{-10pt}
\subsubsection{Effectiveness of Anomaly Detector}

This part analyzes the effectiveness of AD of \sysname in detecting adversarial attacks. First, Fig.~\ref{fig:adv_ks_roc_b} provides the ROC curves with AUROC scores of \sysname for different attacks on the CIFAR-10 dataset under the white-box setting. The plots show that \sysname shows consistently high AUROC scores (0.90 to 0.99) with a very low FPR against most attacks (except Square), making it a reasonable defense for practical settings. 
Moreover,  Table~\ref{tab:auroadversarial_baseline} provides a comprehensive analysis of the effectiveness of \sysname and contrasts with the baselines in detecting adversarial attacks in terms of AUROC scores under both white-box and black-box settings.

The left panel (white-box) of the table shows the performance of the closest baselines where 
 Manda~\cite{wang2022manda} generally struggles against most of the attacks, and MagNet~\cite{meng2017magnet} and Artifacts~\cite{feinman2017detecting} demonstrate reasonable defense only against some of them. Contrarily, consistently high AUROC scores of \sysname show it is highly effective in distinguishing between benign and malicious instances across all attacks and datasets under the white-box setting.

However, under black-box attacks, as demonstrated in the right panel of the table, all baseline methods mostly fail (low AUROC scores) against all of these attacks. Nevertheless, \sysname still remains highly resilient against such attacks. Thus, even if black-box attacks cannot directly compromise the target model's performance, they still leave detectable traces within the input data, which \sysname can effectively leverage. 
{Overall, \sysname achieves average AUROC scores of 0.932 in white-box settings and 0.875 in black-box settings. 
In comparison, MagNet has  0.655 and 0.612, Artifacts has 0.653 and 0.600, and Manda has 0.556 and  0.573, in white-box and black-box settings, respectively.}


\begin{wrapfigure}{r}{0.50\textwidth}
    \centering
    \includegraphics[width=0.995\linewidth]{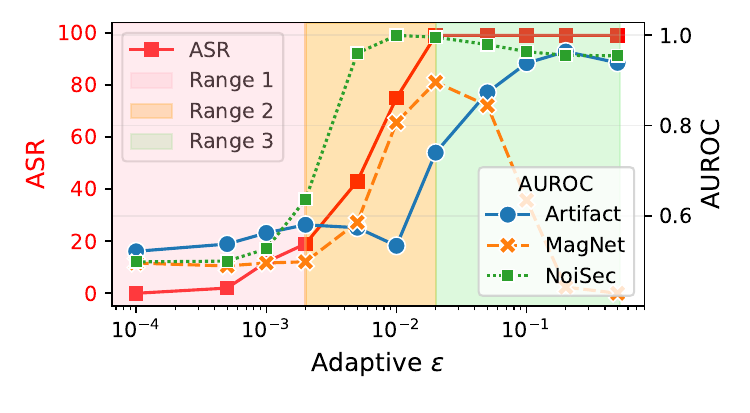}
    \vspace{-30pt}
    \caption{Performance evaluation of \sysname under adaptive attacks where the attacker can adjust the attack strength to avoid detection.}
    \vspace{-10pt}
    \label{fig:adaptive}
    \vspace{-5pt}
\end{wrapfigure}

\subsubsection{Adaptive Adversarial Attacks}
\label{sec:adaptive-attack}

Lastly, we analyze the robustness of {\sysname} against an adaptive adversary who can adjust perturbation strength $\epsilon$ to balance stealth and attack effectiveness. This evaluation uses a representative BIM attack on the CIFAR-10 dataset, considering a range $\epsilon$ from 0.0001 to 0.50. 
{Fig.~\ref{fig:adaptive} shows, for \(\epsilon < 0.002\) (Range 1: high stealth, low effectiveness), ASR remains below 20\%. At $0.002 \leq \epsilon < 0.02$ (Range 2: moderate stealth, moderate effectiveness), ASR increases, reaching 100\% by \(\epsilon = 0.02\). Beyond this \(\epsilon > 0.02\), (Range 3: low stealth, high effectiveness), ASR remains 100\%, showing the stealth-effectiveness trade-off.
Fig.~\ref{fig:adaptive}} also presents the AUROC scores of various detectors for these attacks across the defined ranges. \sysname demonstrates consistent robustness in both ranges 2 and 3, mostly with an AUROC score higher than 0.90. In comparison, MagNet is slightly effective, primarily at the boundary between ranges 1 and 2, while Artifacts performs well only in the latter part of range 3, where the attack stealthiness is very low. These findings highlight \sysname as the only detector capable of maintaining reliable performance across varying levels of attack strength, making it a comprehensive defense against adaptive adversarial threats.


\subsection{Effectiveness against Backdoor Attacks}

\subsubsection{Attack Implementation Results}

Fig.~\ref{fig:backdoor-samples} (in Appendix~\ref{sec:backdoor_samples}) shows the samples with different backdoor triggers. In our implementation of backdoor attacks on the CIFAR-10 dataset, the BadNet attack achieved almost a 100\% ASR but resulted in a drop in benign accuracy to 76.81\%. 
WaNet maintained 
\begin{wrapfigure}{r}{0.60\textwidth}
    \vspace{-20pt}
    \centering
    \subfigure[KS test]{
    \label{fig:backdoor_ks_roc_a}
    \includegraphics[width=0.25\textwidth]{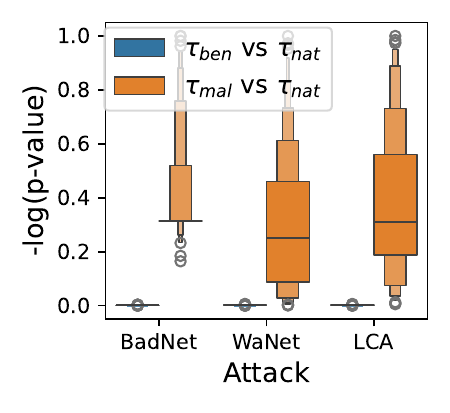}}
    \subfigure[ROC curves]{
    \label{fig:backdoor_ks_roc_b}
    \includegraphics[width=0.30\textwidth]{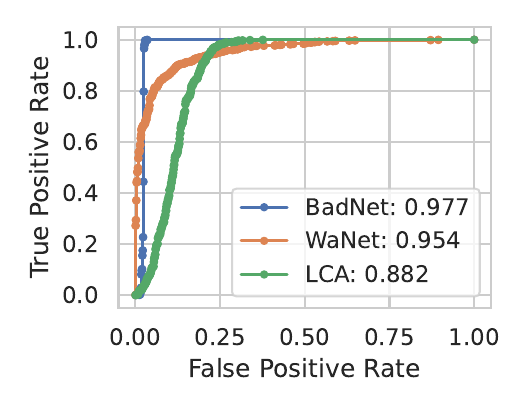}}
    \vspace{-10pt}
    \caption{{\sysname's performance against backdoor attacks.} (a) FE in extracting trigger features from the reconstruction noise and (b) the effectiveness of AD in detecting the existence of the trigger under different backdoor attacks.}
\label{fig:backdoor_ks_roc}
    \vspace{-30pt}
\end{wrapfigure}
strong performance, achieving 92\% benign accuracy and 99\% ASR.  Meanwhile, LCA also maintained 92\% benign accuracy but had a lower ASR of 78\%. 

\vspace{-5pt}
\subsubsection{Effectiveness of Feature Extractor}
\label{effec_feat_rep_badnet}

This analysis evaluates the efficacy of FE in capturing learned trigger features under the backdoor attacks. Similar to Section~\ref{sec:effec_feat_rep_adv}, we compare the feature distributions of reconstructed benign noise ($\tau_{ben}$) and reconstructed backdoor trigger ($\tau_{mal}$) against reconstructed natural noise ($\tau_{nat}$) using the KS test. 
Fig.~\ref{fig:backdoor_ks_roc_a} presents the KS test results for the backdoor attacks. For all the attacks, $\tau_{mal}$ exhibit distinct distributions from $\tau_{nat}$, characterized by higher $-log(p-value)$. 
On the other hand, $\tau_{nat}$ and $\tau_{ben}$ generally share similar distributions and possess lower $-log(p-values)$ values in their KS test. Such a finding further highlights FE's capability to reveal if an input has a backdoor trigger on it. 
This result supports our hypothesis that adversarial and backdoor attacks share common traits that \sysname exploits to design a unified defense mechanism.

\setlength{\tabcolsep}{2pt}
\begin{wraptable}{r}{0.55\linewidth} 
    \vspace{-25pt}
    \centering
    \small
    \caption{Baseline comparison regarding \\AUROC scores against backdoor attacks.}
    \label{tab:baseline-backdoor}
    \resizebox{0.50\textwidth}{!}{%
    \begin{tabular}{|c|c|c|c|}
        \hline
        Defense $\downarrow$ \, Attack $\rightarrow$ & BadNet & LCA & WaNet \\
        \hline
        IBD-PSC  & 0.93 & 0.73 & \textbf{0.99} \\
        SCALE-UP & 0.95 & 0.81 & 0.85 \\\hline
        \sysname &\textbf{0.98} & \textbf{0.88}& 0.95\\
        \hline 
    \end{tabular}
    }
    \vspace{-10pt}
\end{wraptable}

\vspace{-10pt}
\subsubsection{Performance of Anomaly Detector}

In this analysis, we evaluate the effectiveness of AD of \sysname in detecting various backdoor attacks. 
Fig.~\ref{fig:backdoor_ks_roc_b} shows the ROC curve, including the AUROC scores, of \sysname against different backdoor attacks. It is evident from the figure that \sysname is highly effective in detecting backdoor-triggered samples, particularly against the BadNet and WaNet attacks, with AUROC scores of 0.977 and 0.954, respectively, and a very low FPR for both. For the LCA attack, \sysname shows reasonable performance, as this type of attack is generally more challenging to detect.

Table~\ref{tab:baseline-backdoor} compares the AUROC scores of different backdoor defenses, e.g., IBD-PSC~\cite{hou2024ibd} and SCALE-UP~\cite{guo2023scale}. Across all attacks, \sysname consistently outperforms or competes with existing defenses. For the BadNet attack, \sysname achieves the highest score ($0.97$), surpassing IBD-PSC ($0.93$) and SCALE-UP ($0.95$), demonstrating its ability to effectively detect fundamental backdoor threats. Against the LCA attacks, \sysname significantly outperforms the other methods with an AUROC of $0.88$. While IBD-PSC performs marginally better for the WaNet attack ($0.99$ vs. $0.95$), \sysname remains competitive. 
Thus, \sysname’s robustness and superior performance make it a strong candidate for defending against both simple and complex backdoor attacks.

\section{Related Work}
\label{sec:relatedworks}

Adversarial and backdoor attacks on ML models, particularly deep neural networks, have become an area of intense research in recent years. 

\textbf{{Adversarial Attack Detection.~}}
Initial attempts at detecting adversarial attacks focused on statistical methods. Feinman et al. \cite{feinman2017detecting} introduced a technique leveraging Bayesian uncertainty estimates and kernel density to detect adversarial examples. This method was among the first to use statistical properties for adversarial detection.
Several approaches tailor detection mechanisms to specific models or datasets. Metzen et al. \cite{metzen2017detecting} proposed augmenting neural networks with small sub-networks that specialize in identifying adversarial perturbations. This approach allows for model-specific fine-tuning of detection capabilities.
Ensemble methods have also shown promise. Pang et al. \cite{pang2018towards} proposed a method combining multiple weak detectors to improve robustness against adversarial attacks. Similarly, MagNet \cite{meng2017magnet} employs a reformer network to adjust input data and a detector network to identify adversarial examples.
Some research has explored statistical and feature-based methods for adversarial detection, such as statistical tests on the distributions of network activations~\cite{grosse2017statistical}, feature-squeezing technique~\cite{xu2017feature}, etc. 
LiBRe \cite{deng2021libre} used Bayesian neural networks to estimate uncertainty for detecting out-of-distribution adversarial samples.

\textbf{{Backdoor Attack Detection.~}}
Detecting backdoors mostly involves reverse-engineering potential triggers that cause misclassification, assuming these triggers are significantly smaller compared to benign triggers. 
This method relies on efficient reverse engineering techniques and anomaly detection to distinguish original triggers from benign ones \cite{wang2019neural}.
Alternative approaches include distribution-based defenses that model the entire trigger distribution using generative adversarial networks to better capture and eliminate triggers \cite{qiao2019defending}. Additionally, model diagnosis methods assess model behavior with unique inputs to detect anomalies indicative of backdoors, employing techniques like one-pixel signatures \cite{huang2020one} and meta neural trojan detection pipelines \cite{xu2021detecting}. These strategies collectively aim to enhance the resilience of models against backdoor attacks \cite{huang2019neuroninspect}.
Another defense is to eliminate the trigger from the input data. Complete input sanitization uses autoencoder-based reconstruction methods to ensure trigger-free inputs without labeled training data, albeit at a significant computational cost \cite{liu2017neural}. 
While all these defenses are mostly devised for specific attack types, \sysname bridges that gap and provides a unified defense just utilizing the noise.

\section{Conclusion}
\label{sec:conclusion}
ML systems have become increasingly vulnerable to adversarial and backdoor attacks, necessitating robust security measures. In this paper, we introduce \sysname, a detection method that only relies on noise to defend against such threats. \sysname is a novel reconstruction-based detector that isolates noise from test inputs, extracts malicious features, and utilizes them to identify malicious inputs. 
Our comprehensive evaluation of \sysname across a diverse range of datasets and attacks demonstrates its superior performance in detecting both adversarial and backdoor attacks. \sysname consistently outperforms state-of-the-art baselines, achieving average AUROC scores of 0.932 against white-box and 0.875 against black-box adversarial attacks. Notably, against backdoor attacks, \sysname attains an average AUROC of 0.937 on the CIFAR-10 dataset. These results underscore \sysname's potential as a unified, robust, and effective defense mechanism for real-world ML applications. While \sysname reveals a potential avenue for ML defense, it can also work in conjunction with the sample-based defense and further augment detection performance.

\vspace{-10pt}
\subsubsection*{Acknowledgements.} This work was supported in part by the Office of Naval Research under grants N00014-24-1-2730 and N00014-19-1-2621, the National Science Foundation under grants 2235232 and 2312447, and a fellowship from the Amazon-Virginia Tech Initiative for Efficient and Robust Machine Learning.

\bibliographystyle{splncs04}
\bibliography{reference}

\appendix

\section{Model Architectures}
\label{sec:autoencoder}

Tables~\ref{table:model_properties} and~\ref{table:autoencoder_properties} provide an overview of the classification models and detailed descriptions of the autoencoders used across different datasets, respectively.

\begin{table}[h!]
\vspace{-20pt}
\caption{Classification models' details for different datasets}
\label{table:model_properties}
\centering
\scriptsize
\begin{tabularx}{0.989\textwidth}{|c|c|c|c|c|c|c|}
    \hline
    \textbf{Dataset} & \textbf{Model Type} & \textbf{Network} & \textbf{Conv Channels} & \textbf{Flat Dim} & \textbf{Feat Dim} & \textbf{Out Dim} \\
    \hline
    \multirow{2}{*}{\shortstack{F-MNIST}} & Target & 3 Conv, 3 FC &  1$\rightarrow$64 & 1600 & 128 & 10 \\
                                                    & Surrogate & 2 Conv, 2 FC & 1$\rightarrow$64 & 9216 & 128 & 10 \\
    \hline
    \multirow{2}{*}{\shortstack{CIFAR-10\\\& GTSRB}} & Target & 6 Conv, 2 FC & 3 $\rightarrow$ 128 & 2048 & 256 & 10/43 \\
                                                    & Surrogate & 4 Conv, 2 FC & 3 $\rightarrow$ 32 & 2048 & 256 & 10/43 \\
    \hline
    \multirow{2}{*}{\shortstack{Speech \&\\Med-MNIST}} & Target & 10 Conv, 2 FC & 3 $\rightarrow$ 512 & 2048 & 256 & 35/2 \\
                                                 & Surrogate & 10 Conv, 2 FC  & 3 $\rightarrow$ 128 & 512 & 256 & 35/2 \\
    \hline
    \multirow{2}{*}{\shortstack{Activity}} & Target & 10 Conv, 2 FC & 3 $\rightarrow$ 256 & 1792 & 512 & 7 \\
                                                 & Surrogate & 10 Conv, 2 FC  & 3 $\rightarrow$ 128 & 896 & 512 & 7 \\
    \hline
\end{tabularx}
\end{table}
\vspace{-25pt}
\begin{table}[h!]
\centering
\scriptsize
\vspace{-15pt}
\caption{Autoencoder models' details for different datasets}
\label{table:autoencoder_properties}
\begin{tabularx}{0.70\textwidth}{|X|m{3.0cm}|m{1.00cm}|m{1.0cm}|}
    \hline
    \textbf{Dataset} & \textbf{Architecture} & \textbf{Noise\newline Std} & \textbf{Latent Dim} \\
    \hline
    F-MNIST & 6 Conv, 2 FC, 6 Deconv & 0.50 & 256 \\
    \hline
    CIFAR-10 \& GTSRB & 6 Conv, 2 FC, 6 Deconv& 0.10 & 1024 \\
     \hline
    Speech \& Med-MNIST  & 6 Conv, 2 FC,  6 Deconv  & 0.20 & 256 \\
     \hline
    Activity  & 6 Conv, 2 FC,  6 Deconv  & 0.05 & 1024 \\
    \hline
\end{tabularx}
\vspace{-25pt}
\end{table}

\section{Adversarial Attack Implementation Results}

\begin{figure*}[]
    \vspace{-25pt}
    \centering
    \includegraphics[width=0.995\linewidth]{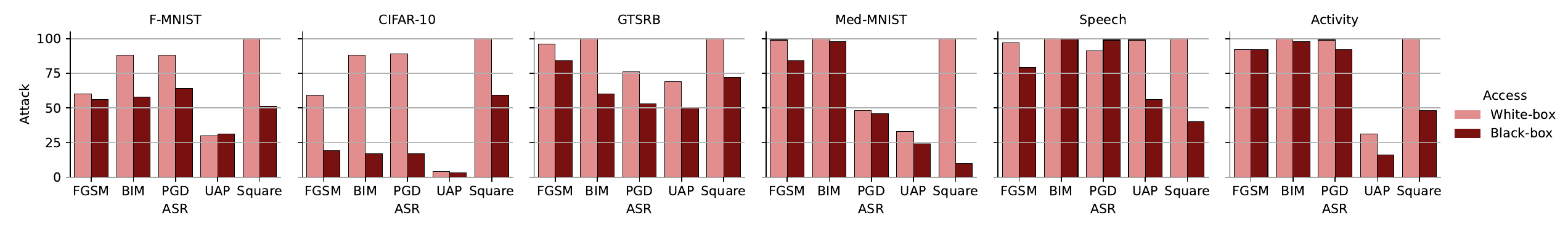}
    \vspace{-10pt}
    \caption{ASR of different attacks across different datasets under both white-box and black-box settings.}
    \vspace{-5pt}
    \label{fig:attack-asr-adv}
\end{figure*}

Fig.~\ref{fig:attack-asr-adv} illustrates the ASR of various attacks across multiple datasets for both white-box and black-box scenarios. In white-box scenarios, the attacks consistently achieve high success rates across all datasets, with many methods reaching ASRs of 80\% or higher,  
showcasing their effectiveness when the model parameters are fully accessible.
Conversely, the performance of black-box attacks presents a stark contrast, demonstrating significantly lower ASR across the same datasets. This decline highlights the inherent challenges that models face under practical, real-world conditions without full access to the models' underlying parameters. For instance, while some black-box attacks show high ASR, 
the overall ASR is considerably diminished compared to their white-box counterparts.
Moreover, regardless of the success of the attacks, either in the white-box or black-box settings, all such attempts need to be detected by the defensive mechanism. 

\section{Detection of Optimization-based Adversarial Attacks}
Optimization-based adversarial attacks, such as the \textit{JSMA}~\cite{papernot2016limitations} and \textit{C\&W}~\cite{carlini2017towards}, involve significant computational overhead due to their reliance on run-time optimization processes, making them less practical in real-world scenarios. Therefore, we primarily focus on the more efficient attack strategies mentioned above. Nevertheless, we also evaluate these optimization-based attacks on the CIFAR-10 dataset to demonstrate the broad applicability of our approach.
\begin{wrapfigure}{r}{0.5\linewidth} 
    \centering
    \vspace{-15pt}
    \includegraphics[width=0.45\textwidth]{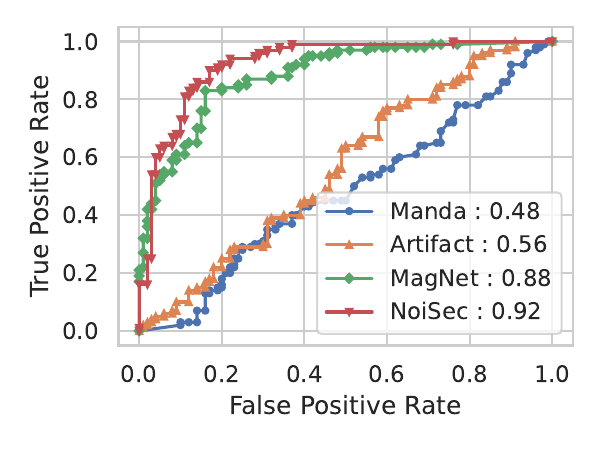}
    \vspace{-5pt}
    \caption{Performance evaluation of \sysname against the optimization-based adversarial attacks on CIFAR-10 dataset.}
    \vspace{-35pt}
    \label{fig:opt_attack}
\end{wrapfigure}

Fig.~\ref{fig:opt_attack} presents the ROC curves and AUROC scores for different detectors against the white-box C\&W attack. As shown, \sysname exhibits high effectiveness with an AUROC score of 0.92, outperforming the closest baseline, MagNet, which achieves an AUROC score of 0.88. 
Moreover, \sysname maintains a low FPR while achieving a high TPR. These results highlight \sysname's robustness against a wide range of adversarial attacks, including gradient-based, optimization-based, and query-based attacks.

\begin{wrapfigure}{r}{0.5\linewidth} 
    \vspace{-25pt}
    \centering
    \includegraphics[width=0.95\linewidth]{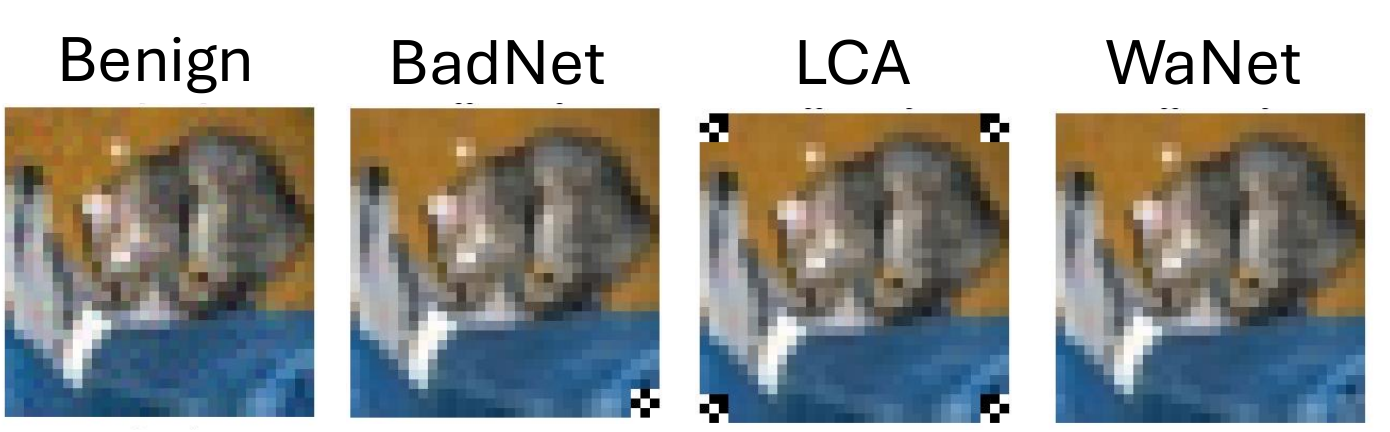}
    \vspace{-5pt}
    \caption{Backdoor triggered samples}
    \label{fig:backdoor-samples}
\end{wrapfigure}

\section{Backdoor Triggered Samples}
\label{sec:backdoor_samples}
Fig.~\ref{fig:backdoor-samples} illustrates backdoor-triggered samples from various backdoor attacks on the CIFAR-10 dataset. Unlike BadNet and LCA, which use visible patterns as triggers, WaNet employs a highly stealthy trigger that mimics natural noise.

\end{document}